\documentclass{article}

\usepackage[preprint]{neurips_2024}

\usepackage[utf8]{inputenc} 
\usepackage[T1]{fontenc}    
\usepackage{hyperref}       
\usepackage{url}            
\usepackage{booktabs}       
\usepackage{amsfonts}       
\usepackage{nicefrac}       
\usepackage{microtype}      
\usepackage{xcolor}         

\usepackage{amsmath}
\usepackage{graphicx}
\usepackage{makecell}

\usepackage{amssymb}

\title{Matching Query Image Against Selected NeRF Feature for Efficient and Scalable Localization}

\author{
 Huaiji Zhou \\
 Binjiang Institute, \ Zhejiang University \\
 \And
  Bing Wang \\
  The Hong Kong Polytechnic University \\
  \AND
  Changhao Chen* \\
  National University of Defense Technology \\
}

\begin{document}

\maketitle

\begin{abstract}
 Neural implicit representations such as NeRF have revolutionized 3D scene representation with photo-realistic quality. However, existing methods for visual localization within NeRF representations suffer from inefficiency and scalability issues, particularly in large-scale environments. This work proposes MatLoc-NeRF, a novel matching-based localization framework using selected NeRF features. It addresses efficiency by employing a learnable feature selection mechanism that identifies informative NeRF features for matching with query images. This eliminates the need for all NeRF features or additional descriptors, leading to faster and more accurate pose estimation. To tackle large-scale scenes, MatLoc-NeRF utilizes a pose-aware scene partitioning strategy. It ensures that only the most relevant NeRF sub-block generates key features for a specific pose. Additionally, scene segmentation and a place predictor provide fast coarse initial pose estimation. Evaluations on public large-scale datasets demonstrate that MatLoc-NeRF achieves superior efficiency and accuracy compared to existing NeRF-based localization methods.
\end{abstract}

\section{Introduction}


Visual localization \cite{localization, loc_survey} has a broad range of applications, including mixed reality \cite{ar}, robotic navigation \cite{robot}, and autonomous driving \cite{auto}. In essence, a visual localization system produces the 6-DoF camera pose of a given query image relative to a pre-built map. This process involves representing the scene as a geometric map, establishing correspondences between key features in the query image and the map, and solving the Perspective-n-Point (PnP) problem to derive the camera pose.

In recent years, neural rendering has garnered significant attention in the realm of mapping and 3D reconstruction. Neural Radiance Fields (NeRF) \cite{nerf} and its variants \cite{mip_nerf, nerfplus} can reconstruct novel views from a given camera pose, transforming camera positions and orientations into the colors and densities of volume-rendered images. Subsequent advancements have further enhanced their robustness \cite{nerf_w} and speed \cite{ngp, baking_nerf}. Today, neural rendering techniques are sophisticated enough to model large-scale scenes, including entire cities. Approaches like Block-NeRF \cite{block} and Mega-NeRF \cite{mega} partition scenes into overlapping spherical sub-regions, each represented by an individual NeRF network. Grid-NeRF \cite{grid_nerf} uses local grid features to guide NeRF training in large-scale scenes, while Switch-NeRF \cite{switch_nerf} decomposes the scene and represents it with a mixture of NeRFs. These approaches can be seen as modeling entire scenes into an implicit neural map.

Compared to the extensive research on using NeRF for scene reconstruction, the use of trained NeRF models for localization, particularly in large scenes, remains relatively underexplored.
NeRF-based localization methods generally fall into two categories. The first category includes inverted NeRF (iNeRF) \cite{inerf} and subsequent works \cite{latitude, nerf_refinement}, which tackle the inverse problem of NeRF: starting from an initial pose estimate, these methods use gradient descent to minimize the residual between pixels rendered from a NeRF and pixels in an observed image. The second category involves matching image features against the neural map using PnP. There are two main ways to obtain features from the neural map. Methods such as FVLocNeRF \cite{FVLocNeRFF} first render RGB images from the map and then extract features using an off-the-shelf RGB feature extractor \cite{orbs}. More recent approaches, such as Crossfire \cite{crossfire}, train neural maps to directly generate image features for matching.




Though NeRF-based localization shows promise in certain scenarios, it faces challenges in efficiency and scalability in real-world applications. For efficiency, iNeRF and its variants require hundreds of iterations for pose refinement, taking approximately 20 seconds to obtain poses, which is impractical for real-time use. Existing matching-based localization methods often incorporate additional features or parameters beyond the NeRF network, making them computationally expensive and memory-intensive.
Regarding scalability, NeRF-based localization methods often struggle with large scenes. Previous neural rendering approaches \cite{mega, block} typically partition NeRF into grid subblocks, which is inefficient for localization tasks, as one image may correspond to several blocks, necessitating a merging strategy. Moreover, previous methods rely on the initial pose being close to the ground truth to deliver fast and accurate results. This highlights the need for a fast coarse pose estimation step before employing NeRF models for pose refinement.
 
To solve the above two concerns, we propose \textbf{MatLoc-NeRF}, a novel \textbf{Mat}ching based \textbf{Loc}alization framework using selected \textbf{NeRF} features, that is efficient and scalable for large-scale visual localization. Our MatLoc-NeRF leverages a learnable feature selection mechanism to efficiently utilize informative NeRF features for matching with query images. This approach surpasses traditional methods that rely on all NeRF features or introduce additional descriptors. By strategically selecting a subset of relevant intermediate features from the NeRF network, MatLoc-NeRF achieves accurate pose estimation while maintaining computational efficiency. This is achieved by matching the selected NeRF features with features extracted from the query image. The 2D-3D correspondences obtained through this process enable pose recovery via the PnP algorithm. 
To handle large-scale scenes effectively, MatLoc-NeRF employs a pose-aware scene partitioning strategy. This strategy ensures that only the most relevant NeRF sub-block generates key features for a specific pose. Additionally, the scene is segmented into co-visible regions based on camera viewpoints. A place predictor then provides the localization model with a coarse initial estimate. Extensive evaluations on public large-scale datasets demonstrate that MatLoc-NeRF achieves superior efficiency and accuracy compared to existing NeRF-based localization methods.


In summary, our main contributions are as follows:
\begin{itemize}
  \item We propose MatLoc-NeRF, a novel matching based localization framework using selected NeRF features, that is efficient and scalable for visual localization. Our feature selection simultaneously minimizes the computation cost and localization error.
  \item We propose a pose-aware scene partitioning strategy that facilitates efficient handling of large-scale scenes by ensuring each pose corresponds to a single NeRF subnetwork.
  \item We introduce automatic clustering for fast pose estimation that enables rapid coarse localization via classification.
\end{itemize}


\section{Related Work}
\textbf{Large-scale Neural Radiance Fields:}
The seminal work of Neural Radiance Field (NeRF) \cite{nerf} shows the possibility of photo-realistic novel view synthesis by representing the scene with an implicit neural network. Subsequent works improve its performance \cite{mip_nerf, nerfplus}, robustness \cite{nerf_w}, and efficiency \cite{ngp, baking_nerf}. Recent works propose methods for scaling the NeRF representation to large-scale scenes. Block-NeRF \cite{block} and Mega-NeRF \cite{mega} partition the scenes into spherical sub-regions and represent each sub-region with an individual NeRF network. Grid-NeRF \cite{grid_nerf} uses local grid features to guide NeRF training in the large-scale scene. Switch-NeRF \cite{switch_nerf} learns to decompose the large-scale scene and represent the scene with a mixture of NeRFs. UE4-NeRF \cite{NEURIPS2023_b94d8b03} uses polygon meshes to improve efficiency in the large-scale scenes.

\textbf{NeRF based Localization:}
There are generally two categories of work on NeRF-based localization. The first category focuses on accurate pose estimation in pre-trained NeRFs. iNerf \cite{inerf} and its extended variants \cite{latitude, nerf_refinement} fall into this category. Latitude \cite{latitude} extends the problem to city-scale scenes with blocks of NeRFs and employs a coarse-to-fine estimation approach. Crossfire \cite{crossfire}, NeRF-Loc \cite{nerf_loc}, and NeFeS \cite{nerf_refinement} enhance robust outdoor localization by incorporating additional features such as neural feature fields or point clouds alongside NeRFs.
The second category considers NeRF-based localization in the context of SLAM or image reconstruction with noisy poses. NeRF-- \cite{nerfmm} and BARF \cite{barf} are pioneering works that jointly optimize camera poses and NeRF. Subsequent research has improved this joint optimization using hash encoding \cite{curriculuminerf} and generalizable NeRF \cite{generalized_dbarf}. iMAP \cite{imap}, NeRF-SLAM \cite{nerf_slam}, and Nice-SLAM \cite{nice_slam} utilize NeRFs as the sole map representation, employing them for both camera tracking and reconstruction.

Our work belongs to the first category of NeRF-based localization methods and employs a matching-based approach. It leverages a pre-trained NeRF to generate scene features and match them with query images for camera pose estimation. FVLocNeRFF \cite{FVLocNeRFF} uses NeRF to render images and depths from initial pose guesses, extracting and matching ORB \cite{orbs} features to iteratively refine the pose. Crossfire \cite{crossfire} adds a feature head to NeRF for direct matching, while NeRF-Loc \cite{nerf_loc} combines NeRFs with back-projected image features.
Unlike these methods, our approach does not use additional features. Instead, we select useful features within the pre-trained NeRF network for matching and localization. We are also the first to optimize matching-based localization from NeRF representations for large-scale scenes.


\section{Methodology}
\label{sec:Methodology}

\begin{figure}[htbp]
  \centering
  \includegraphics[width=0.85\textwidth]{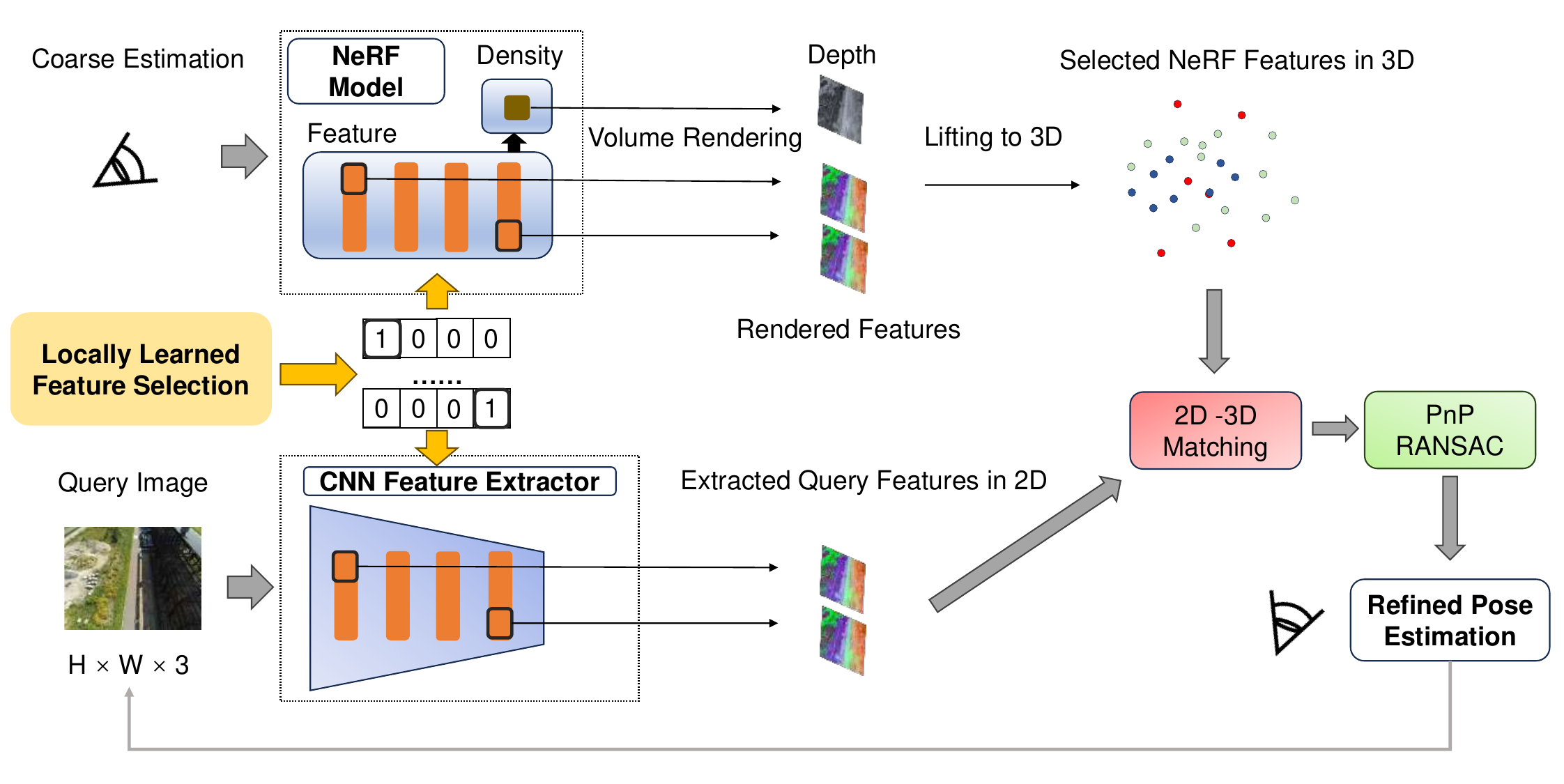}
  \caption{Our MatLoc-NeRF, a novel matching-based NeRF localization framework learns to select the useful NeRF features for 2D-3D matching with the query image, efficient and scalable localization is achieved by solving the PnP problem with these matches.}
  \label{figure: overall}
\end{figure}

Figure \ref{figure: overall} illustrates the overall framework of our proposed MatLoc-NeRF framework. 
We first establish 2D-3D correspondences between the query image and NeRF's intermediate features for localization (Section 3.1.1). A learnable feature selection process then minimizes the number of features used, enhancing efficiency (Section 3.1.2). Finally, scalable localization is achieved through our introduced scene and camera pose partitioning strategies (Section 3.2).

\textbf{Preliminaries of NeRF:} 
The Neural Radiance Field (NeRF) encodes a scene using multi-layer perceptrons (MLP). For each point in the scene, the density \(\sigma\) and color \(\mathbf{c}\) are evaluated by the mapping function \(F_{\mathbf{\Theta}}(x, y, z, \theta, \phi) \rightarrow (\sigma, \mathbf{c})\), where \(\mathbf{\Theta}\) are the parameters of MLPs, and \((x, y, z)\) and \((\theta, \phi)\) denotes the 3D coordinates and 2D viewing direction, respectively. The RGB pixel colors \(\mathbf{C}(\mathbf{r} )\) are obtained through volume rendering along its camera ray \(\boldsymbol{r}(\mathrm{t})=\boldsymbol{o}+\mathrm{t} \boldsymbol{d}\). The integral of volume rendering is approximated using numerical quadrature \cite{quad}:
\begin{equation}
\begin{gathered}
\mathbf{C}(\mathbf{r} ; \Theta)=\sum_k T_k\left(1-\exp \left(-\sigma_{k}\left(t_{k+1}-t_k\right)\right)\right) \mathbf{c}_k = \sum_k w_k \mathbf{c}_k \\
\quad \text { where } \quad T_k=\exp \left(-\sum_{k^{\prime}<k} \sigma_{k^{\prime}}\left(t_{k^{\prime}+1}-t_{k^{\prime}}\right)\right)
\end{gathered}
\end{equation}
\(\sigma_{k}\) and \(\mathbf{c}_k\) are the density and colors at the k th sampling interval \([t_k, t_{k+1}]\) along the camera rays.
Given a set of training images with corresponding camera poses, the parameters of the NeRF network can be optimized by minimizing the differences between rendered pixel colors and ground truths.

\subsection{Matching based Localization Using Selected NeRF Features}

\subsubsection{Matching Query Image against NeRF Features}
Unlike prior works \cite{crossfire, nerf_loc} that introduce additional features or parameters, MatLoc-NeRF focuses on maximizing efficiency and scalability without modification to the original NeRF model. We achieve this by exploring the potential of NeRF's intermediate features for localization. Thus, we propose a novel learning strategy to select features most relevant for matching the query image to NeRF features during localization, eliminating the need for additional computational costs.

The intermediate features of NeRF are the values of intermediate layers in the NeRF network. Specifically, given a 3D point \((x, y, z)\), a positional-encoding layer \(F_\text{pos}\) and stacks of MLP layers \(F_\text{mlp}\) are applied to \((x, y, z)\) sequentially, producing the intermediate features \(\mathbf{f_{pos}} = F_\text{pos}(x, y, z)\) and \(\mathbf{f_{mlp}} = F_\text{mlp}(F_\text{pos}(x, y, z)\). NeRF then uses these intermediate features to decode the density and direction-dependent colors of this point: \(\sigma = F_{\sigma}(F_\text{mlp}(F_\text{pos}(x, y, z)))\), \(\mathbf{c} = F_{c}(F_\text{mlp}(F_\text{pos}(x, y, z)), \theta, \phi)\). Compared to rendering the scene colors with \(F_{c}\) or creating additional features, obtaining these intermediate features requires less computation. 



MatLoc-NeRF leverages intermediate features extracted from the NeRF network for efficient localization. For a given query image, a pre-trained CNN projects its features into the same domain as the intermediate NeRF features, enabling effective matching. We then initialize a pose for the query image using coarse estimation techniques. Given this pose, intermediate features can be rendered from the NeRF network using the same volumetric rendering integral as described in Equation (1):
\begin{equation}
\mathbf{F}(\mathbf{r}) = \sum_k w_k \mathbf{f}_k
\end{equation}
Similarly, We could obtain the depth of the rendered intermediate features with:
\begin{equation}
\text{U}(\mathbf{r}) = \sum_k w_k u_k
\end{equation}
where \(u_k\) is the distance of the sampled point from the camera and \(\mathbf{f}_k\) is the combined intermediate features of \(\mathbf{f_{pos}}\) and \(\mathbf{f_{mlp}}\) from NeRF at the sampled point. \(w_k\) is the same weighting from accumulation as defined in Equation (1). With depth estimation \(\text{U}(\mathbf{r})\), rendered intermediate features \(\mathbf{F}(\mathbf{r})\) are lifted into 3D space, forming a point cloud of intermediate features around the scene surface.

For establishing correspondence between intermediate features in the 3D space and 2D features in the query image. We construct a similarity matrix \(\mathbf{M} \in \mathbb{R}^{(H \times W)\times(H \times W)}\), where \((H, W)\) is the query image size. We then identify mutual nearest neighbors, forming dense pairs of 2D-3D matches between the 2D features of the query image and the 3D features of the scene. With these dense pairs of 2D-3D correspondences, a Perspective-n-Point (PnP) solver can be employed to determine the relative camera pose \(\mathbf{P} \in SE(3)\) between the query image and the initialized pose. The PnP solver operates within a RANSAC loop to handle outliers in the matches. Finally, a refined camera pose estimation is obtained by transforming the initialized camera pose with the calculated relative pose.

\subsubsection{Learning to Select Intermediate Features from NeRF Model}


While NeRF's intermediate features are optimized for color and density prediction, their suitability for direct feature matching in localization is uncertain. Joint training of NeRF for both rendering and localization is possible, but it would be computationally expensive and require careful balancing of loss functions within the limited capacity of the NeRF network.

To address this challenge, our MatLoc-NeRF utilizes a learnable binary selection matrix \(\mathbf{s} \in \mathbb{R}^D\) to identify and leverage the most effective intermediate features for localization. This approach minimizes the number of features used, improving efficiency. The selection process minimizes the sum of differences between corresponding features from 2D-3D pairs in the query image and rendered scene. This formulation ensures the selection of the minimal number of informative features required for the scene. The feature selection process is formulated as a binary integer programming problem:

\begin{align} 
\begin{split}
\text{Minimize} \sum_{(i,j)\in Pairs} \frac{\sum_{d=1}^{D} \mathbf{s}_d (F_d^{2D}[i] - F_d^{3D}[j])}{\sum_{d=1}^{D} \mathbf{s}_d}
\end{split}\\
\begin{split}
\quad \text{s.t.} \quad \quad \quad \mathbf{s}_d = 0, 1
\end{split}\\
\begin{split}
\quad \quad \quad \quad \quad \ 0 \leq \sum_{d=1}^{D} \mathbf{s}_d \leq N_s
\end{split}
\end{align}
where the binary selection variable \(s_d\) determines whether the $d$-th element in the \(\text{D}\)-dimensional features in 2D (\(F_d^{2D}\)) and 3D (\(F_d^{3D}\)) are selected for matching, \(Pairs\) contains a set of index pairs \((i,j)\) denoting the ground truth matches, and the constraint ensures only less than \(N_s\) of the intermediate features are selected for localization. This eliminates feature redundancy and further improves efficiency by requiring less feature computation for localization.
To generate "ground truth matches" for learning the selections, we sample an image from the training database and initialize a pose estimation near it. We then generate the corresponding features with NeRF and establish dense 2D-3D matches. Incorrect matches above a certain threshold are filtered using the known relative pose between two poses. The remaining "ground truth matches" are added to the training set \(Pairs\) for selection learning. 

NeRF's capability to generate 2D-3D image pairs for any pose near the query image allows for ample training data to represent various cases of initialized poses. By dividing the scene into sub-regions and optimizing feature selection for each region, we enable data parallelism and allow for the selection of different features when new scene blocks with different domains are added.

\subsection{Scaling NeRF-based Localization to Large-scale Scenes}
Previous NeRF-based localization approaches \cite{crossfire, nerf_loc} primarily target small scenes represented by a single NeRF. Applying a single NeRF to large-scale scenes becomes inefficient and impractical. Current scene partitioning methods \cite{mega, block} for rendering large scenes rely on sample point locations. However, if points from the same pose fall into different sub-regions, rendering requires multiple sub-NeRFs, hindering efficient feature generation for localization. Additionally, existing methods often function as pose refinement tools, achieving high accuracy only when the initial pose is close to the ground truth. To address these challenges in large-scale scenes, we introduce two novel partitioning strategies:
1) Efficiency-aware scene partitioning: optimizes the NeRF re-localization pipeline for faster feature generation. 2) Efficient pose partitioning for coarse estimation: improves the efficiency of initial pose estimation. This two-pronged approach ensures efficient and accurate localization in large-scale environments.

\subsubsection{Pose-Aware Scene Partition Strategy}
Previous works such as Mega-NeRF and Block-NeRF addressed large-scale scene rendering by partitioning NeRF model into sub-regions of overlapping spheres. Each sub-region is then represented by a smaller NeRF network (sub-NeRF). They leverage nearby sub-NeRFs for training on images and rendering based on camera pose. While this improves data parallelism and rendering efficiency compared to a single network, it may not be optimal for localization tasks. We propose an alternative scene partitioning strategy specifically designed for efficient localization. Our method prioritizes task-specific efficiency, leading to faster and more robust localization performance.

Given a camera pose \(\mathbf{P} \in SE(3)\), the number of sub-NeRFs used for rendering features or color of the 3D scene: \(\text{Num\_NeRF}(\mathbf{P})\), is directly proportional to the model parameters stored in memory. A large \(\text{Num\_NeRF}(\mathbf{P})\) also reduces the degree of parallelism and computational efficiency, as the sample batch of 3D points will be divided into sub-batches and processed by different sub-networks. While partitioning the scene into spheres does not guarantee the smallest overall \(\text{Num\_NeRF}\), our goal is to minimize the average number of sub-NeRF used for all possible poses without compensating the network capacity of each sub-NeRF, enabling more efficient feature rendering for matching.
We begin with deriving \(\text{Num\_NeRF}\) for a given camera pose \(\mathbf{P}\). During the volumetric rendering process as defined in Equation (1), the ray sampler maps the camera pose \(\mathbf{P}\) to a batch of sample point coordinates along its camera rays in 3D space:
\begin{equation}
\{\text{sample}_1(\mathbf{P}), \text{sample}_2(\mathbf{P}), ... \ \text{sample}_n(\mathbf{P})\} \rightarrow \{(x_1, y_1, z_1), (x_2, y_2, z_2), ... \ (x_n, y_n, z_n)\} 
\end{equation}
Depending on the scene partition strategy, each 3D coordinate is allocated to a sub-region with sub-NeRF: \(\text{NeRF\_ID}(x, y, z) = \text{allocator}(x, y, z)\). The number of NeRFs used for a given pose $P$ is the total unique number of used sub-NeRFs after allocation:
\begin{equation}
\text{Num\_NeRF}(\mathbf{P}) = \text{Unique}\{\text{NeRF\_ID}(x_1, y_1, z_1), ... \ \text{NeRF\_ID}(x_n, y_n, z_n)\}
\end{equation}
Our goal is to design a partitioning strategy with the allocator that yields a lower average \(\text{Num\_NeRF}(\mathbf{P})\) over the pose distribution in the scene. Although it is non-trivial to solve for the optimal partition considering all possible partitions. We can force the allocator to allocate all samples from a pose \(\mathbf{P}\) to the same \(\text{NeRF\_ID}\). Using this allocation, only one sub-NeRF is used in every camera pose:
\begin{equation}
\text{allocator}(\text{sample}_i(\mathbf{P})) = \text{NeRF\_ID}_{\mathbf{P}} \quad\rightarrow \quad \text{Num\_NeRF}(\mathbf{P}) \triangleq 1
\end{equation}
There are generally more camera poses and fewer sub-NeRFs. With \text{Num\_NeRF} minimized, our goal becomes allocating camera poses to sub-NeRFs such that their capacities in feature and color rendering are maximized. Results in Table ~\ref{tab: scene size per pose and capacity} suggest the rendering quality of NeRF decreases when the coordinates of sample points are more dispersed across the scene. This finding aligns with spatial allocation strategies in prior works, which compact each sub-region into a spherical space. Thus, the partition strategy should allocate poses to NeRFs such that, within each NeRF, sample points of allocated poses are compact in space and have minimal spatial variance. 

For each camera pose, sample point coordinates along its camera rays can be viewed as a point cloud. Our partition strategy clusters these pose-wise point clouds so that the spatial variance of sample point coordinates in each cluster is minimized, and camera poses in each cluster are allocated to the same NeRF. A direct approach to clustering these point clouds is the K-Means algorithm with Chamfer point cloud distance as the distance metric. However, computing point cloud distances is expensive, and K-means has a complexity directly proportional to the distance computations. We adopt a more effective strategy by converting pose-wise point clouds into occupancies of 3D voxel grids. The intersection of voxel grid occupancies is used as the distance metric for clustering.



\subsubsection{Automatic Clustering for Fast Coarse Pose Estimation}

\begin{figure}[htbp]

  \centering
  \includegraphics[width=0.9\textwidth]{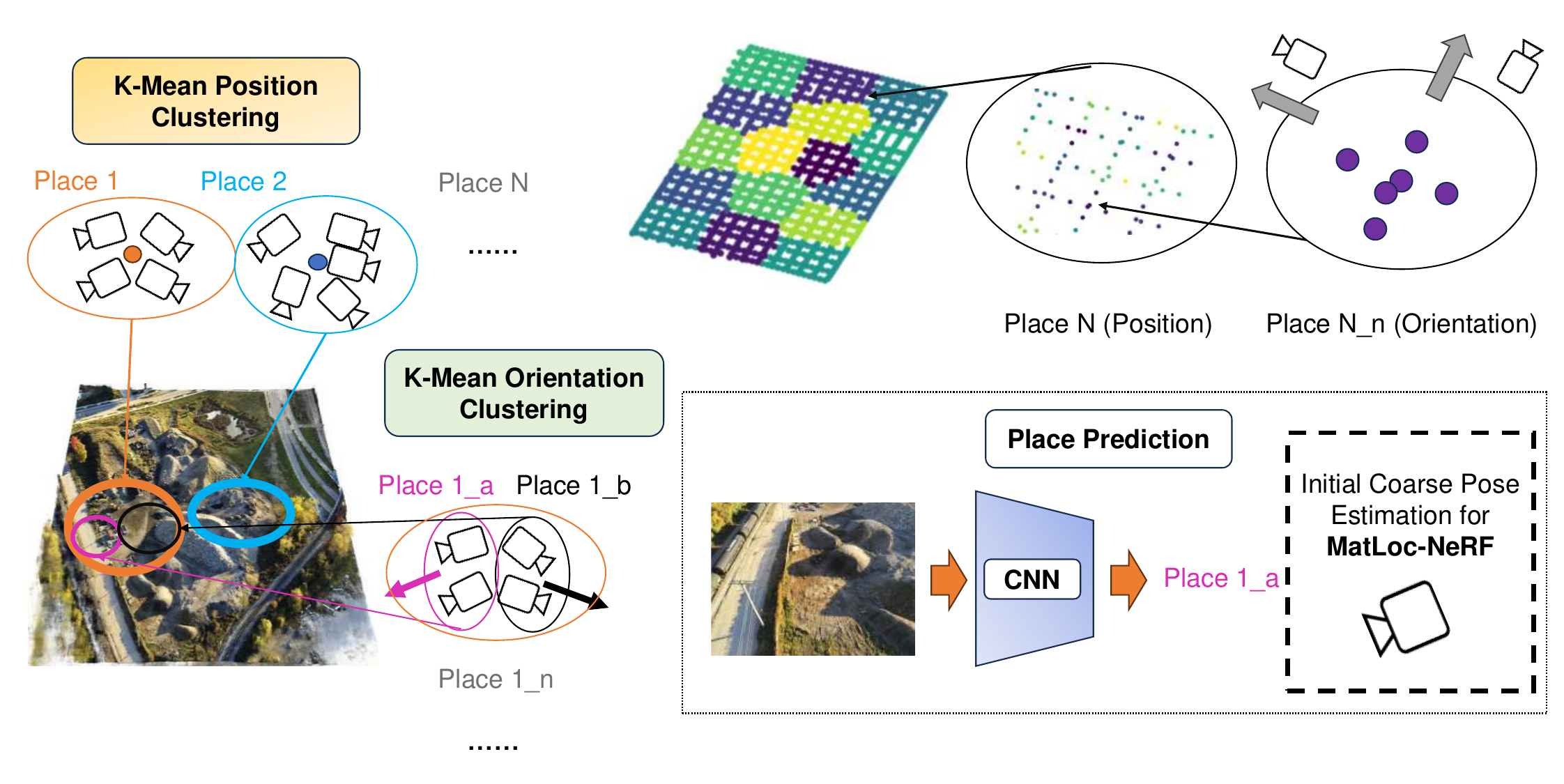}
  \caption{Our coarse pose estimation module uses 2-stage K-Means cluttering to identify candidate locations within the large-scale scene. a lightweight CNN is trained to predict the scene location corresponding to the query image, providing an initial pose for NeRF-based matching. }
  \label{figure: cluster}
\end{figure}

NeRF-based localization methods rely on an initial pose estimate close to the true location. This initial estimate is then refined for greater accuracy. However, in large-scale scenes, random initialization can lead to slow convergence or complete failure. This is because randomly chosen poses might not share any common view with the ground truth (lack of co-visibility).

To address this challenge, we propose a hierarchical localization approach using pose groups. Traditional methods for hierarchical re-localization retrieve images from a database that share a view with the query image. These images, often referred to as "places," are then used along with nearby 3D map points for pose refinement. However, maintaining a separate image database becomes memory-intensive for NeRF-based methods that represent the scene implicitly with neural networks.
Our solution leverages "pose groups" instead of "places." Each pose group represents a cluster of camera poses with similar positions and orientations (Figure \ref{figure: cluster}). Since cameras within a group share a similar view of the scene, they inherently exhibit co-visibility. This allows us to achieve efficient pose refinement without the need for an additional image database.

We employ K-Means clustering and perform two rounds of clustering sequentially: spatial clustering followed by orientation clustering. First, camera poses are spatially clustered based on camera translations as the distance metric. The spatially similar camera pose groups are then further divided into sub-groups with similar orientations, using angles between the camera orientations as the distance metric. 
Our pose partitioning scheme automatically groups all poses in the scene into co-visible places. The coarse pose estimation can be reformulated as a place prediction problem of K-Means clustered "place". For a query image, we extract a d-dimensional feature using a ResNet34 backbone and train a CNN for predicting the "place" of the pose cluster in the large-scale scene using Additive Angular Margin Loss (ArcFace) \cite{deng2019arcface}, which learns individual class embeddings to separate a large number of classes.


\section{Experiments}
\subsection{Experimental Setup}
\textbf{Datasets:} We conducted experiments on two large-scale scenarios of a widely-adopted public dataset, i.e. Mill-19 \cite{mega} dataset. The Rubble scene comprises over 1600 training images and 21 testing images, all with resolutions of 4068\(\times\)3456, and corresponding 6-DoF camera poses. We also tested our method on the Building scene with over 1900 images and corresponding camera poses that capture urban environments. We downscaled all the images by a factor of 4 for more efficient scene reconstruction and localization. 

\textbf{Evaluaton Metrics:} 
For the primary localization task, we report the median rotation (°) and translation errors (\(m\)) on both datasets. Additionally, we employed PSNR, SSIM, and LPIPS metrics to assess rendering quality during ablation studies focused on NeRF's rendering capabilities.

\textbf{Implementation Details:}
We implemented our models using PyTorch \cite{pytorch} and the NerfStudio \cite{nerfstudio} framework. The Gekko \cite{gekko} library facilitated solving the non-linear integer programming problem. We built upon Nerfacto, a NeRF model known for its efficient and accurate trade-off, with two key modifications: 1) Multi-stage lightweight sampling: This enhancement improves efficiency by employing a multi-stage sampling approach. 2) Multi-resolution hash encoding of local features: This modification accelerates both training and inference by utilizing a multi-resolution hash encoding scheme for local features.
We trained each model with a batch size of 4096 and optimized the parameters using the Adam \cite{kingma2014adam} optimizer.
The learning rate is set to \(1\text{e}^{-2}\), and an exponential decay scheduler is used with a final learning rate of \(1\text{e}^{-4}\). 

\subsection{NeRF Localization Performance in Large-scale Scenes} 
We compare the results of our method with other NeRF-based localization methods in the large-scale scenes (500\(\times\)250 \(m^2\) ) captured by Rubble and Building \cite{mega} datasets. Our coarse pose estimation module (Section 3.2.2) provided initial pose estimates for all methods.
While iterative methods like iNeRF \cite{inerf}, which refines the pose by minimizing the difference between a rendered image and the query image, often fail to converge precisely on these large-scale scenes. This may be due to the complex loss landscape of NeRF at this scale. Similarly, the curriculum iNeRF pose update approach employed by the \cite{curriculuminerf} variant of iNeRF also exhibits convergence issues on both datasets.
Matching-based methods, such as the one proposed by FVNeRFLoc \cite{FVLocNeRFF} (Rendered RGB+Descriptor), which relies on rendered image matching with descriptor key points, achieve localization based on 3D geometry. We use SIFT \cite{sift} as the descriptor for its efficient implementation. However, their accuracy is highly dependent on rendering quality. 
Our method outperforms these existing approaches by achieving the lowest rotation and translation errors in both scenarios as shown in Table \ref{tab:pose results}.

\begin{table}[hbt!]
  \caption{
    The quantitative results of NeRF-based localization methods on the Rubble and Building scenarios of Mill-19 Dataset.
  }
  \label{tab:pose results}
  \centering
  \begin{tabular}{cc|cc|cc} 
    \toprule
    \multicolumn{2}{c|}{} & \multicolumn{2}{|c|}{Rubble}& \multicolumn{2}{|c}{Building}\\
    \midrule
    Pose Estimation Method & Category & Trans.(m)$\downarrow$ & Rot.(°)$\downarrow$& Trans.(m)$\downarrow$ & Rot.(°)$\downarrow$\\
    \midrule
    iNeRF \cite{inerf} & Backprop. & 5.6 &  10.3 & 3.4 & 5.8\\
    Curriculum iNeRF \cite{curriculuminerf}& Backprop. & 2.8 &  6.4 & 3.2 & 7.5\\
    Rendered RGB \cite{FVLocNeRFF} + SIFT \cite{sift} & Matching & 4.1 &  1.4 & 2.1 & \textbf{1.2}\\
    \midrule
    MatLoc-NeRF (Ours) & Matching & \textbf{1.7} &  \textbf{0.8} & \textbf{1.8} & \textbf{1.2}\\
  \bottomrule
  \end{tabular}
\end{table}

\subsection{Efficiency Analysis}
\textbf{Localization Efficiency:} 
Compared to iterative methods like iNeRF and its variants, which require hundreds of iterations for pose updates through forward and backward passes in the NeRF network, our method achieves significantly faster localization. This is because these iterative approaches necessitate repeated network evaluations for refinement.
Matching-based methods, like FVNeRFLoc, offer some improvement by requiring only a single forward pass to extract color and features for matching. However, our method surpasses even this approach in terms of speed. We leverage intermediate features directly for matching, eliminating the need for full RGB rendering, resulting in the lowest latency for localization (Table \ref{tab: efficiency 0}).

\begin{table}[hbt!]
  \caption{
    A comparison of  efficiency for NeRF-based localization methods
  }
  \label{tab: efficiency 0}
  \centering
  \begin{tabular}{c|c|c|c} 
    \toprule
    Method & 
    {\makecell[c]{NeRF \\ Iterations}} &
    {\makecell[c]{Localization \\ Time}}\\
    \midrule
    iNeRF \cite{inerf} & 300 & 18 s\\
    Curriculum iNeRF \cite{curriculuminerf} & 200 & 12 s\\
    Rendered RGB \cite{FVLocNeRFF} + SIFT \cite{sift} & 1 & 1.9 s\\
    \midrule
    MatLoc-NeRF & \textbf{1} & \textbf{1.4 s}\\
  \bottomrule
  \end{tabular}
\end{table}

\textbf{Impact of Feature Selection on Matching:}
The Matching-based localization performance depends on the quality of the chosen intermediate features. Our method, detailed in Section 3.1.2, incorporates a learned selection process to identify the most effective features for accurate localization. This selection process is formulated with constraints, guaranteeing that only a limited set of features (denoted by \(N_s\)) are used for matching. This focus on a reduced feature set directly translates to faster matching times as detailed in Table \ref{tab: efficiency 1}.

\begin{table}[hbt!]
  \caption{
    A comparison of  efficiency for MatLoc-NeRF using different intermediate features for matching and localization
  }
  \label{tab: efficiency 1}
  \centering
  \begin{tabular}{c|c|c|c|c} 
    \toprule
    {\makecell[c]{Intermediate Feature \\ Used for Localization}} &
    {\makecell[c]{Feature \\ Dimension}} &
    {\makecell[c]{Matching \\ Time}} &
    {\makecell[c]{Trans.(m)$\downarrow$}} &
    {\makecell[c]{Rot.(°)$\downarrow$}} \\
    \midrule
    All Intermediate & 47 & 0.87 s  & 1.73 &  0.84\\
    Positional Intermediate & 32  & 0.87 s & 1.75 & 1.03\\
    MLP Intermediate & 15 & 0.79 s & 1.77& 0.95 \\
    Learned Selection (\(N_s\)=10)  & 10 & 0.77 s &  1.73 & \textbf{0.78} \\
    Learned Selection (\(N_s\)=5)  & 5 & \textbf{0.74 s} &  \textbf{1.70} & 0.82 \\
  \bottomrule
  \end{tabular}
\end{table}

\textbf{Efficiency in Coarse Pose Estimation:} While MatLoc-NeRF's coarse pose estimation step aims to predict a co-visible location for initializing the pose refinement stage in NeRF matching localization, both our method and existing pose regression techniques achieve sufficient accuracy for successful matching. However, our method accomplishes this with significantly lower latency – nearly half the speed of previous approaches as shown in Table \ref{tab: coarse pose results}.

\begin{table}[hbt!]
  \caption{
    A Comparison of inference time for coarse pose estimation
  }
  \label{tab: coarse pose results}
  \centering
  \begin{tabular}{c|c|c} 
    \toprule
    Method & 
    {\makecell[c]{Coarse Estimation \\ Time}}\\
    \midrule
    Pose Regression \cite{latitude} & 0.061 s \\
    MatLoc-NeRF (Coarse ) & \textbf{0.027 s}\\
  \bottomrule
  \end{tabular}
\end{table}


\subsection{Ablation Study on Scene Partition Strategies}
\textbf{Partition Efficiency} 
This section explores the efficiency of different scene partition strategies through ablation studies. Our proposed method assigns all sample points captured from a single camera pose to the same sub-NeRF, resulting in a fixed number of sub-NeRFs (denoted as \(\text{Num\_NeRF(P)}\)) regardless of the scene complexity. Experiment results in Table \ref{tab: partition num} demonstrate that this allocation strategy leads to faster feature generation for localization tasks compared to the grid block partition methods employed by Block-NeRF and Mega-NeRF.

\begin{table}[hbt!]
  \caption{
    The quantitative results for the impact of \(\text{Num\_NeRF(P)}\) on efficiency.
  }
  \label{tab: partition num}
  \centering
  \begin{tabular}{c|c|c} 
    \toprule
    Partition Method &  \(( \text{Num\_NeRF(P)})\) & {\makecell[c]{Feature Generation Time}}\\
    \midrule
    Mega-NeRF (Grid Partition) \cite{mega} & 2.1 & 0.83 s\\
    MatLoc-NeRF (Pose-aware Partition) & \textbf{1} & \textbf{0.67 s}\\
  \bottomrule
  \end{tabular}
\end{table}

Our clustering-based partition strategy goes beyond efficiency in feature generation. It also promotes compact scene representations within each sub-NeRF. This is achieved by ensuring that camera poses assigned to the same sub-NeRF capture a tightly packed region of the scene volume. As evidenced by the results in Table \ref{tab: scene size per pose and capacity}, this approach leads to maximized efficiency, likely due to the reduced scene complexity within each sub-NeRF.

\begin{table}[hbt!]
  \caption{
    Impacts of average scene size covered per camera pose on a single NeRF network's capacity in rendering and re-localization.
  }
  \label{tab: scene size per pose and capacity}
  \centering
  \begin{tabular}{c|c|c|c|c|c} 
    \toprule
    {\makecell[c]{Sampling Space of\\ NeRF Sub-block(\(m^3\))}} & {\makecell[c]{Translation \\ Error (\(m\)) \(\downarrow\)}} & {\makecell[c]{Rotation \\ Error (°) \(\downarrow\)}}  & PSNR\(\uparrow\)  & SSIM\(\uparrow\)  & LPIPS\(\downarrow\) \\
    \midrule
    70  & 1.97 & 0.28 & 17.72 & 0.4472 & 0.6384 \\
    50  & 1.73 & 0.28 & 18.59 & 0.4663 & 0.6061 \\
    35 & \textbf{1.67} & \textbf{0.18} & \textbf{21.32} & \textbf{0.5642}& \textbf{0.5171}\\
  \bottomrule
  \end{tabular}
\end{table}

\textbf{Impact of Sub-NeRF number on Localization:} 
We investigated the effect of the number of sub-NeRFs on localization accuracy. This involved partitioning the scene into sub-NeRFs, allocating 50, 100, and 200 camera poses to each sub-NeRF configuration. As expected, decreasing the number of sub-NeRFs led to a decline in rendering quality due to the limited number of sub-NeRFs covering the entire scene. Interestingly, we observed a slight improvement in localization accuracy with lower rendering quality. This might be attributed to the removal of redundant high-resolution features that could potentially hinder the localization process. To achieve a balance between efficient localization and maintaining high-quality rendering, we opted for a configuration of 150 poses per sub-NeRF.


\section{Conclusion}
This work introduces MatLoc-NeRF, an efficient and scalable NeRF-based localization framework for large-scale environments. By using a learned feature selection process within the existing NeRF network, MatLoc-NeRF enables fast and accurate matching without relying on additional features.
Our novel scene partition strategy accelerates feature generation and rendering, while a clustering-based approach for coarse pose estimation enhances efficiency. MatLoc-NeRF offers valuable insights into localizing large-scale scenarios with implicit neural fields, paving the way for faster and more robust localization in complex environments.

\bibliographystyle{plain}
\bibliography{reference}

\end{document}